%% file: sample-base.tex
\documentclass[letterpaper, 10 pt, conference]{ieeeconf}

\IEEEoverridecommandlockouts                              % This command is only needed if 
\usepackage{cite}
\usepackage{amsmath,amssymb,amsfonts, mathtools}
\usepackage{algorithm}
\usepackage{breqn}
\usepackage{algpseudocode}
\usepackage{graphicx}
\usepackage{textcomp}
\usepackage{xcolor}
\usepackage{bmpsize}
\usepackage{caption}
\usepackage{amsmath}
\usepackage{amsfonts}
\usepackage{amssymb}
\usepackage{multirow}
\usepackage{accents}
\usepackage{hyperref}
\usepackage{academicons}

\usepackage{booktabs, multirow} % for borders and merged ranges
\usepackage{soul}% for underlines
\usepackage{changepage,threeparttable} % for wide tables
\usepackage{authblk}

\usepackage{flushend}

\begin{document}

\title{\LARGE \bf 
Object Goal Navigation using Data Regularized Q-Learning}

\author{Nandiraju Gireesh$^1$, D. A. Sasi Kiran$^{1}$, Snehasis Banerjee$^2$, Mohan Sridharan$^3$ \\ Brojeshwar Bhowmick$^2$, Madhava Krishna}
\affil[1]{Robotics Research Center, IIIT Hyderabad, India}
\affil[2]{TCS Research, Tata Consultancy Services, India}
\affil[3]{Intelligent Robotics Lab, University of Birmingham, UK}

\renewcommand\Authands{ and }

\maketitle

\begin{abstract}
Object Goal Navigation requires a robot to find and navigate to an instance of a target object class in a previously unseen environment. Our framework incrementally builds a semantic map of the environment over time, and then repeatedly selects a long-term goal ('where to go') based on the semantic map to locate the target object instance. Long-term goal selection is formulated as a vision-based deep reinforcement learning problem. Specifically, an Encoder Network is trained to extract high-level features from a semantic map and select a long-term goal. In addition, we incorporate data augmentation and Q-function regularization to make the long-term goal selection more effective. We report experimental results using the photo-realistic Gibson benchmark dataset in the AI Habitat 3D simulation environment to demonstrate substantial performance improvement on standard measures in comparison with a state of the art data-driven baseline. 
%achieves 63.7\% Success and 31.3\% SPL, a 10\% and 12\% relative improvement over the state of the art method \cite{semexp}. 
\end{abstract}
\begin{keywords}
Deep Reinforcement Learning, Data Augmentation, Q-value Regularization, Object Goal Navigation.
%Reinforcement Learning, Embodied Agents, Visual Navigation.
\end{keywords}

\input{sections/1_introduction}
\input{sections/2_relatedwork}
\input{sections/3_approach}
\input{sections/4_setup}

\input{sections/5_results}
\input{sections/6_conclusion}

\bibliographystyle{IEEEtran}
\bibliography{sample-base}

\end{document}

%% file: sections/1_introduction.tex
\section{Introduction}\label{section:introduction}
Object-goal navigation (ObjectNav) is a key task in robotics, which requires a robot to navigate to an instance of a target object class in a previously unseen environment~\cite{objectnav}. This is a challenging task because the robot needs to address multiple problems such as understanding the current scene from sensor input (e.g., RGB-D observation of the scene), computing a suitable path to the likely location of an instance of the target object, and navigating reliably to the desired location. Conventional methods for the ObjectNav task usually require comprehensive domain knowledge, e.g., map of the environment, which is not feasible in many practical domains. Although navigation methods based on end-to-end learning methods have resulted in promising results on specific datasets, they incur high computational costs and tend to generalize poorly to previously unseen scenes. % some papers using RL state that generalization to unseen scenes is good like 
% \begin{figure}[ht] 
%         \centering
%         \includegraphics[width=\linewidth]{sections/figures/Object Nav.png}
%         \caption{\textbf{Object Goal Navigation: } Illustrative example in which the objective is to navigate to an instance of a target object class such as `dining table'. The robot is initialized at a random location in the environment. At each timestep, it receives RGBD (image) observations and pose readings and takes navigational actions to reach the target object instance.}
%         \label{task}
% \end{figure}

Modular reinforcement learning (RL)-based methods for ObjectNav have emerged as a strong competitor to end-to-end RL methods with better sample efficiency, generalization to new scenes, and transfer from simulation to the real world. They rely on separate modules for mapping, exploration, and navigation. The state of the art modular RL method for ObjectNav, Goal-Oriented Semantic Exploration (SemExp)~\cite{semexp}, has low overall accuracy due to failure cases that can be attributed to poor selection of \textit{long-term goals}, i.e., intermediate regions to explore for the target object. 

% \begin{figure*}[t] 
%         \centering
%         \includegraphics[width=0.9\textwidth]{sections/figures/Model Overview.png}
%         \caption{Our framework consists of three main components. The \textit{Semantic Mapping} module uses RGB-D observations and odometry pose readings to build an allocentric semantic map of the local world. Our \textit{Encoder Network} extracts high-level features from the estimated semantic map. These features are used to train an \textit{Actor-critic network} which samples a long-term goal to reach the target object class instance efficiently. Analytical planners are used by the deterministic local policy to compute low-level navigation actions to reach any given long-term goal.}
%         \label{fig1}
% \end{figure*}

This paper poses the `Where to go?' (long-term goal selection) problem as a vision-based RL problem. Deep Reinforcement Learning (DRL) methods represent the state of the art in multiple domains such as video games, robot manipulation, and visual navigation. However, these methods, by themselves, often make it difficult to compute policies that generalize to unseen environments, even when the new environments are similar to those used to compute the policies. Strategies to address this limitation are based on learning better low-dimensional representations using auto-encoders, variational inference, contrastive learning, self-prediction, or data augmentation. Drawing on these insights, we introduce a DRL framework for ObjectNav task, which makes the following contributions:
%as well as regularize the Q-function in the updation process of the encoder, actor and critic networks which is built upon Soft Actor-Critic \cite{sac}. % WRITE AS 2 Sentences
%To summarize our paper makes the following contributions:
\begin{itemize}
    \item Incorporates an \textit{Encoder Network} to learn rich, high-level features from an estimated semantic (domain) map, and an \textit{Actor-Critic Network} that is trained with these features to provide effective long-term goals for the ObjectNav task. 

    \item Incorporates (image) data augmentation in the context of the semantic domain map, and regularizes the Q(value) function in the update process of the encoder, actor, and critic networks in the framework, leading to better generalization and long-term goal selection.

    %  \item We formulate "where to go?" as a vision-based RL algorithm and demonstrate how simple image augmentation on the predicted semantic map helps in a much effective long-term goal selection. We also qualitatively show how our proposed method has a better long-term goal selection.
    % \item We use two simple mechanisms for regularizing the value function which are generally applicable in the context of model-free off-policy RL.
\end{itemize}
We evaluate the framework's capabilities using benchmark (photo-realistic) scenes from the Gibson dataset in the AI Habitat 3D simulation environment~\cite{habitat} and show that our approach provides a 10-12$\%$ relative improvement in established performance measures in comparison with a state of the art baseline for ObjectNav task. Additional qualitative results and supporting material are available online: \url{https://user432.github.io/objnav-drq/}
% method achieves 63.7\% Success and 31.3\% SPL on the ObjectNav task, a relative improvement of 10\% and 12\% over prior state of the art \cite{semexp} respectively.
%\Mohan{it would help to include an illustrative figure of the environment and task here!}

%% file: sections/2_relatedwork.tex
\begin{figure*}[!tb] 
        \centering
        \includegraphics[width=0.9\textwidth]{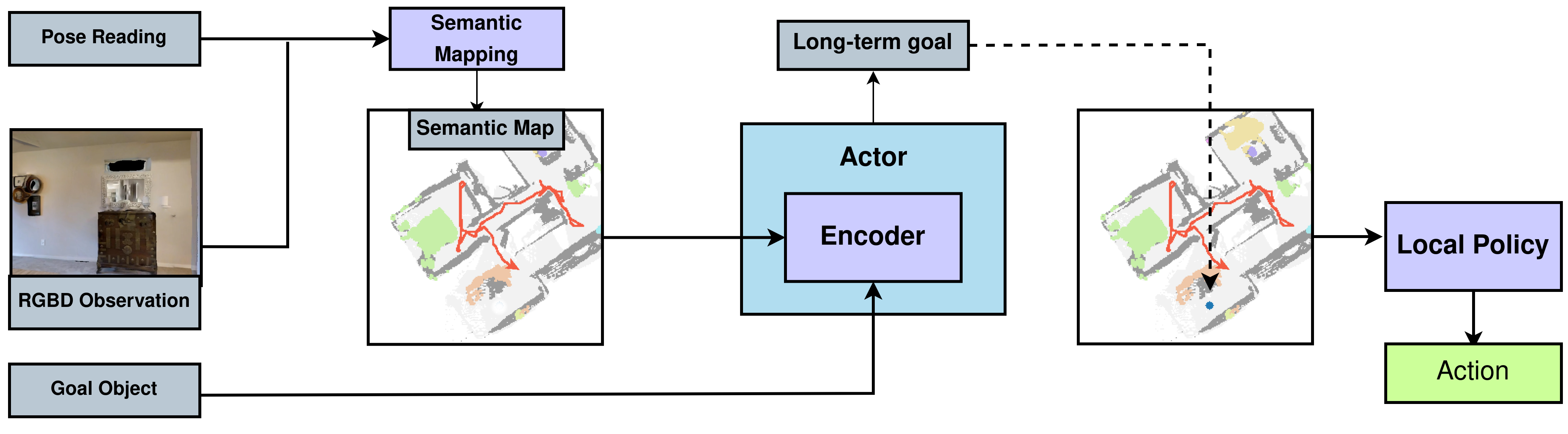}
        \vspace{-0.5em}
        \caption{Our framework consists of three main components. The \textit{Semantic Mapping} module uses RGB-D observations and odometry pose readings to build an allocentric semantic map of the local world. Our \textit{Encoder Network} extracts high-level features from the estimated semantic map. These features are used to train an actor-critic network which samples a long-term goal to reach the target object class instance efficiently. Analytical planners are used by the deterministic \textit{Local Policy} to compute low-level navigation actions to reach any given long-term goal.}
        \label{fig1}
        \vspace{-0.5em}
\end{figure*}

\section{Related Work}\label{related work}
We review related work in Object Goal Navigation and the learning of representations through reinforcement.

\textbf{Object Goal Navigation.} There has been much work on developing learning algorithms for vision-based recognition and navigation. Data-driven deep (reinforcement) learning methods are state of the art for learning to map pixels directly to actions~\cite{cogplan,pix2act}. These methods find it difficult to generalize to previously unseen scenarios since they do not build a representation of the environment. Methods that seek to promote better generalization with such methods construct an allocentric map that encodes semantic priors~\cite{semexp,cogmap,ans}. In this paper, we draw inspiration from this insight to pose the `where to go?' decision as a vision-based deep RL problem, and build an explicit semantic map to set an effective long-term goal for the ObjectNav task.

\textbf{Self-supervised Representation Learning.} Many self-supervised methods have been developed in the computer vision literature to learn representations for different tasks; state of the art methods include contrastive methods~\cite{simclr,moco} as well as predictive methods~\cite{dino,simsiam}. The learned representations have been shown to improve the performance of the corresponding supervised learning system, particularly when the amount of labeled training data available for the associated tasks is limited. The key insight from these methods, which we build on, is that good representations can be acquired from visual input through unsupervised learning or data augmentation, and that the learned representation can significantly improve performance on the associated task.

\textbf{Representation Learning in RL.} There has been a lot work on combining the ideas of self-supervised learning and RL in order to improve sample efficiency and performance. In particular, unsupervised learning of representation in conjunction with data augmentation methods have helped improve the efficiency of RL methods. For example, researchers have used auto-encoders~\cite{autoenc} to improve efficiency of RL for visual information processing tasks~\cite{sac-ae}. Other self-supervised learning methods have helped link state-based and image-based RL, e.g., using contrastive learning~\cite{curl}, self-prediction~\cite{spr}, and augmented data~\cite{drq,rad}. Our work in this paper is inspired by the DrQ system~\cite{drq}, which demonstrated the use of regularization to significantly improve the performance of the Soft Actor Critic method~\cite{sac} when trained on image pixels in the context of tasks from the DeepMind Control (DMC) suite~\cite{dmcontrol}. However, unlike their focus on solving continuous control tasks based on information encoded in image pixels, we pursue a modular approach for better generalization and efficiency, and focus on effectively sampling long-term goals by introducing a learned semantic map in the pipeline.

%% file: sections/3_approach.tex
\section{Problem Description and Approach}
\label{section:approach}
%In this section we describe the ObjectNav task and then introduce our method.
%\subsection{ObjectNav Definition}
The ObjectNav task initializes a robot in a random location in a previously unexplored environment and requires the robot to navigate to the closest instance of a target object class (e.g., chair, bed). More specifically, the robot's inputs at each time step $t$ include $640 \times 480$ RGB-D images ($s_t$), odometer readings $(x, y, \theta)$, and the target object category $o$. %The odometer readings are aggregated over time to obtain the agent’s relative pose $p_t$ (relative to pose at $t = 0$). The agent then executes an action $a_t$ at time $t$. 
The robot has to navigate within $d_s = 1.0m$ of the target object class instance for successful task completion. %The episode terminates when the agent executes stop, or if it exceeds a time budget of $T = 500\ steps$. % is it time or step count upper limit pre-set?

\begin{figure*}[t] 
        \centering
        \includegraphics[width=0.9\textwidth]{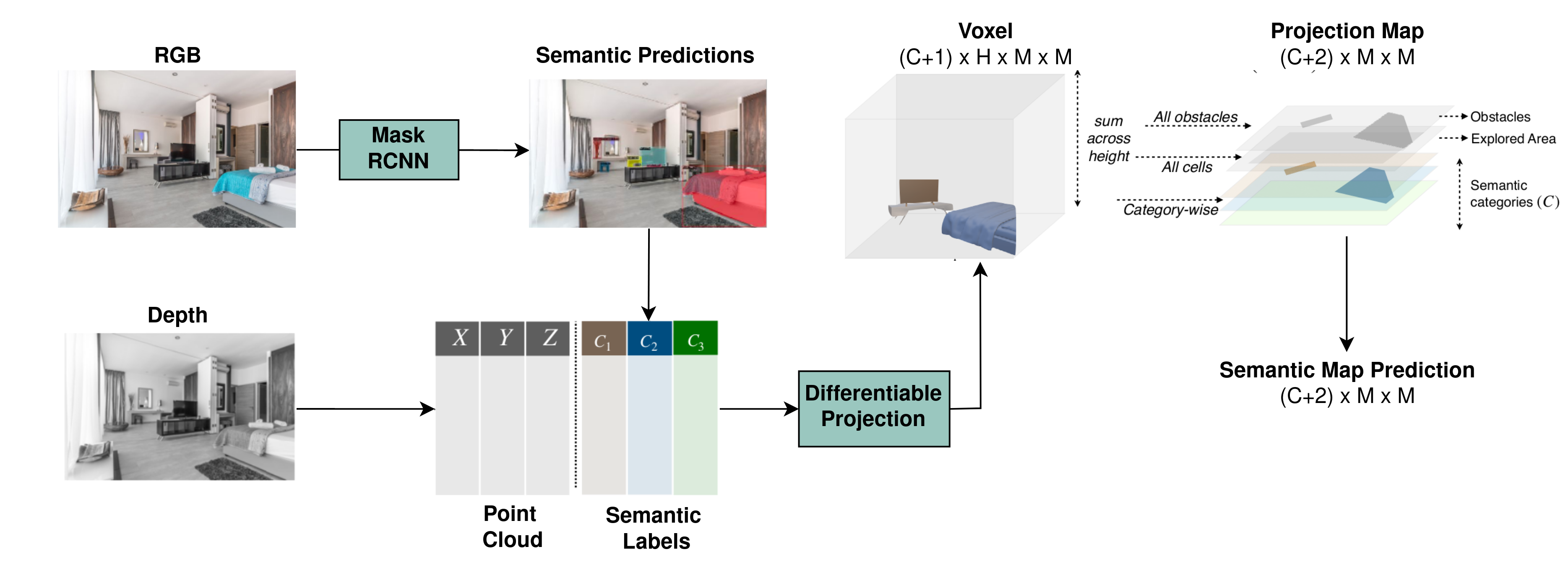}
        \vspace{-0.75em}
        \caption{\textbf{Semantic Mapping.} A sequence of RGB and Depth images are processed through a sequence of operations to produce a top-down Semantic Map.}
        \label{fig2}
        \vspace{-1em}
\end{figure*}

\subsection{Framework Overview}
\label{section:approach-overview}
Figure~\ref{fig1} provides an overview of our framework for the ObjectNav task. It has three key modules: \textit{Semantic Mapping}, \textit{Encoder Network}, and a deterministic \textit{Local Policy}; the novelty is in the adaptation and proposed combination of these modules. The Semantic Mapping module builds a semantic map from the RGB-D images and pose observations (Section~\ref{section:approach-semanticmap}). This map and the object goal are passed to the Encoder Network, which extracts high-level features and sends them to the Actor-Critic Network to select a long-term goal that improves performance on the ObjectNav task (Section~\ref{section:approach-encoder}). An analytical planner is used to compute a deterministic local policy that provides low-level navigational actions to reach the long-term goal (Section~\ref{section:approach-policy}). We describe the individual modules below. %first describe the Semantic Mapping module and then describe our proposed Encoder Network along with it's training procedure.

\subsection{Semantic Mapping}
\label{section:approach-semanticmap}
Figure~\ref{fig2} summarizes the operation of the Semantic Mapping module, which is responsible for constructing and maintaining a semantic metric map $m_t$ and pose of the robot $p_t$. The map $m_t$ is a $K \times M \times M$ matrix where $M\times M$ is the map size. Each element of this spatial map represents a cell of size $25cm^2$ $(5cm \times 5cm)$ in the physical world. The number of channels in the semantic map is $K = C + 2$, where $C$ is the total number of semantic categories. As shown in the top right of Figure~\ref{fig2}, the first two channels represent obstacles and the explored area, and each of the remaining channels represent an object category. For each cell in the map, the channels indicate whether the corresponding location is an obstacle, has been explored, or contains an object of a particular category. The pose $p_t \in R^3$ denotes the $(x, y)$ coordinates and the orientation of the robot at time t. The initial position of the robot is at the map's center, facing East at the start of the episode, i.e., $p_0 = (M/2, M/2, 0.0)$.

The procedure used to create the semantic map is a variant of the process followed by the SemExp method, the  state of the art baseline for ObjectNav task~\cite{semexp}. First, we use a pretrained Mask R-CNN~\cite{maskrcnn} model to estimate the semantic categories from the RGB (image) observations. Next, the depth observations are used to compute point-clouds that are registered in an allocentric coordinate system using the robot’s sequence of poses $(p_0,..., p_t)$. Each point in the point cloud is associated with the estimated semantic categories. A voxel representation is then built using differentiable geometric computations over each point in the point cloud. This voxel representation is converted into a $(C+2)\times M \times M$ semantic map $m_t$. As stated earlier, channels 1 and 2 correspond to obstacles and explored areas, while the remaining channels correspond to the $C$ object categories.

One of the innovations in our framework is the use of image augmentation methods with the semantic map to promote generalization. We explored four such methods:
\begin{itemize}
\item \textbf{Random Shift:} semantic maps of size $240 \times 240$ are padded on each side by four pixels (by repeating boundary pixels) and then randomly cropped back to $208 \times 208$ size.

\item \textbf{Horizontal/Vertical Flip:} this method simply flips the semantic map either horizontally or vertically with probability $0.1$.

\item \textbf{Grayscale:} this method converts our RGB semantic map into a grayscale image.

\item \textbf{Rotate:} this method rotates our semantic map by $r$ degrees, where $r$ is uniformly sampled from $[-5, +5]$.
\end{itemize}
In Section~\ref{section:results}, we discuss the experimental evaluation of these methods and justify the choice of the \textit{random shift} method for data augmentation in our work.

\subsection{Encoder Network}
\label{section:approach-encoder}
The estimated semantic map is passed to our encoder network, along with the robot's current and past locations and the target object class, to learn better representations. \textit{This introduction of an encoder is a key innovation that, as we show later, contributes to a performance improvement in the ObjectNav task}. The encoder extracts high-level features that are used by the robot's actor network to generate a long-term goal, i.e., the region the robot should travel to next to look for an instance of the target object class.

The encoder network architecture comprises of $4$ convolutional layers with $3\times3$ kernels and $32$ channels~\cite{drq}; ReLU activation is applied after each convolution layer. We use stride length of one everywhere. The output of these layers is fed into a single fully-connected layer normalized by LayerNorm~\cite{layernorm}. Finally, we apply the hyperbolic tangent nonlinear transform to the 50-dimensional output of the fully-connected layer. We initialize the weight matrix of the fully-connected and convolutional layers through the orthogonal initialization method~\cite{orthogonal} and set the bias as zero.

The actor-critic network is set up to operate on the output of the encoder network. The actor and critic components have separate encoders, although they share the same weights in the convolution layers. Also, only the optimizer in the critic is allowed to update these weights, e.g., the gradients from the actor do not propagate to the shared convolution layers. We employ the clipped double Q-learning method~\cite{doubleq} for the critic. In this method, each Q-function is parameterized as a three-layer multi-layer perceptron (MLP) with ReLU activations after each layer except the last one. The actor is also a three-layer MLP with ReLU activations; it outputs the mean and covariance for the diagonal Gaussian that represents the policy. The hidden layer's dimension is $1024$ for both the critic and the actor.

As mentioned earlier, we regularize the Q-function in the update process of the relevant networks to promote generalization. Regularizing the Q-function of the critic network, for example, results in different transformations of the same semantic map to have the same Q-function values~\cite{drq}. 

Our framework uses a replay buffer to store all the transition states, including the semantic map, action, reward, next semantic map, goal category, and next goal category. A batch of transitions $(s, a, r, s',g, g')$ are sampled from the replay buffer and used with the augmented semantic map as input to the encoder and actor-critic networks. A long-term goal is sampled once every $25$ timesteps. During training, the decrease in the distance to the nearest target object instance after the completion of the episode is used as reward to revise the parameters of the networks. 
%The specific approach used to update the actor-critic and encoder network parameters is explained further below.
%In our proposed method, in addition to augmenting the semantic map, we regularize the Q-function learned by the critic network so that different transformations of the same semantic map have similar Q-function values as proposed in Kostrikov et al. \cite{drq}.

%We apply random shift augmentation to the semantic map and obtain the augmented semantic map, where the predicted map of size $240 \times 240$ are padded each side by $4$ pixels (by repeating boundary pixels) and then randomly cropped back to $208 \times 208$ size. This procedure is repeated every time a semantic map is sampled from the replay buffer.

Algorithm~\ref{alg:1} describes the steps for updating the actor-critic network. There are two key differences from the standard soft actor-critic update steps. During the update of the critic network, we average the target Q-values over $K=2$ image transformations:
\begin{multline}
    y_i=r_i+\gamma\frac{1}{K}\sum_{k=1}^{K}Q_\theta(f(s'_i,v'_{i,k}),a'_{i,k},g'_i),\\ \text{where}\ a'_{i,k}\sim\pi(\cdot|f(s'_i,v'_{i,k}),g'_i)
\end{multline}
where state $s$ can be treated as some function $f(\cdot)$ with parameters $v$. Also, we then average the Q-function itself over $M=2$ image transformations:
\begin{equation}
    \theta\leftarrow\theta - \lambda_\theta \nabla_\theta \frac{1}{NM}\sum_{i=1,m=1}^{N,M}(Q_\theta(f(s_i,v_{i,m}),a_i,g_i)-y_i)^2
\end{equation}
where $N$ is the sampled batch size. The other update steps remain unchanged. 
%Everything else is the same, and the actor the critic networks are updated in a Soft Actor-Critic way as shown in 

\begin{algorithm}[tb]
\caption{Steps for updating actor-critic network}\label{alg:1}
\begin{algorithmic}
\State \textbf{Hyperparameters:} Total number of environment steps $T$, mini-batch size $N$, learning rate $\lambda$, target network update rate $\tau$, image transformation $f$, number of target $Q$ augmentations $K$, number of $Q$ augmentations $M$.
\For {each timestep = $1...T$}
    \State $a_t \sim \pi(\cdot|s_t,g_t)$
    \State $s'_t \sim p(\cdot|s_t,a_t)$
    \State $D \leftarrow D \cup (s_t,a_t,r_t,s'_t,g_t,g'_t)$
    \State $\textsc{UpdateCritic}(D)$
    \State $\textsc{UpdateActor}(D)$ %\Comment{Data augmentation is used for actor training as well.}
\EndFor
\Procedure{UpdateCritic}{$D$}
    \State ${(s_i, a_i, r_i, s'_i, g_i, g'_i)}^{N}_{i=1} \sim D$ \Comment{Sample a mini batch}
    \For{each $i=1...N$}
        \State $a'_i \sim \pi(\cdot|f(s'_i,v'_{i,k}),g'_i),\ k=1...K$
        \State $\hat{Q_i} = \frac{1}{K} \sum^{K}_{k=1} Q_{\theta'}(f(s'_i,v'_{i,k}),a'_{i,k},g'_i)$
        \State $y_i \leftarrow r_i + \gamma \hat{Q_i}$
    \EndFor
    \State $J_Q (\theta) = \frac{1}{NM} \sum_{i=1,m=1}^{N,M}(Q_\theta(f(s_i,v_{i,m}),a_i,g_i)-y_i)^2$
    \State $\theta \leftarrow \theta - \lambda \nabla_\theta J_Q(\theta)$ \Comment{Update the critic}
    \State $\theta' \leftarrow (1 - \tau)\theta' + \tau \theta$ \Comment{Update the critic target}
\EndProcedure
\Procedure{UpdateActor}{$D$}
    \State ${(s_i, g_i)}^{N}_{i=1} \sim D$ \Comment{Sample a mini batch}
    \State $a_i \sim \pi(\cdot|f(s_i,v_{i,k}),g_i),\ k=1...K$
    \State $J(\phi) = \text{log}\ \pi(a_i|s_i,g_i) -  Q_\theta(f(s_i,v_{i,k}),a_i,g_i)$
    \State $\phi \leftarrow \phi - \lambda \nabla_\phi J(\phi)$ \Comment{Update the actor}
\EndProcedure
\end{algorithmic}
\end{algorithm}

\subsection{Local Policy}
\label{section:approach-policy}
The local policy enables the robot to navigate to the long-term goal $g_t$ sampled by the actor after the actor-critic network is trained. It is obtained using the Fast Marching Method~\cite{fmm}, which computes the shortest path from the present location to the long-term goal using the obstacle channel from the semantic map $m_t$. The robot invokes the local policy to execute deterministic actions and move along the computed shortest path. If a physical robot is performing the ObjectNav task, we could use (for example) a sampling-based motion planner for navigation; we do not do so here because that is beyond the scope of this paper.

%% file: sections/4_setup.tex
\section{Experimental Setup}\label{setup}
We use the Gibson benchmark dataset of scenes in the AI Habitat simulator \cite{habitat} for our experiments. Gibson consists of scenes which are 3D reconstructions of real-world environments. We use the \textit{train} and \textit{val} splits of the \textit{Gibson Tiny} dataset for training and testing respectively.% as the test set is held-out for the online evaluation. 
We do not use the validation set for hyper-parameter tuning. From the train split, we only trained on 10 scenes at a time due to limitations on the computational resources available for use. Instead we repeated the training and evaluation five times. In each such repetition, we randomly sampled $10$ scenes from the Gibson train split dataset, and trained our framework and the baseline for $1\ million$ frames. In each repetition, the trained models were evaluated on the same (standard) Gibson val split (with $5\ scenes$).

As stated earlier, the observation space consisted of the RGB-D images of size $4 \times 640 \times 480$, and the success threshold $d_s=1m$. The maximum episode length was $500$ steps. For the target object categories, we used six object categories which were common between the Gibson dataset and the MS-COCO dataset: `chair', `couch', `potted plant', `bed', `toilet' and `tv' (television).

\begin{table}[tb]
\begin{center}
\caption{Overview of our framework's \textbf{hyperparameters}.}
\label{hyperparameters} 
\begin{tabular}{|c|c|} %change to cc for 2 columns
\hline
\multicolumn{1}{|c|}{\textbf{Parameter}} & \multicolumn{1}{c|}{\textbf{Setting}}\\
\hline
Replay buffer capacity & 40000 \\
Minibatch size & 8 \\
Discount($\gamma$) & 0.99 \\
Optimizer & Adam \\
Critic Learning rate & $10^{-3}$ \\
Critic Q-function soft-update rate ($\tau_Q$) & 0.01 \\
Critic encoder soft-update rate ($\tau_{enc}$) & 0.05 \\
Actor Learning rate & $10^{-3}$ \\
Actor log std-dev bounds & [-20,2] \\
Init temperature & 0.1 \\
Features Dimension & 50 \\
Hidden Dimension & 1024 \\
\hline
\end{tabular}
\end{center}
\end{table}

We experimentally evaluated the following hypotheses about the capabilities of our framework:
\begin{itemize}
    \item[\textbf{H1:}] Our framework provides better performance on the ObjectNav task in comparison with the baselines.% the encoder network performs substantially well when the success rate and SPL are compared with the baseline.
    \item[\textbf{H2:}] Our framework provides better long-term goals than the state of the art DRL baseline. 
\end{itemize}
We evaluated these hypotheses by comparing our method with the following two baselines.
\begin{enumerate}
\item \textbf{Random:} The robot chooses actions randomly from the set of Habitat simulator actions: \textit{move\_forward, turn\_right, turn\_left,} and \textit{stop}.

\item \textbf{Semantic Exploration (SemExp):} state of the art modular DRL method for ObjectNav, which won the ObjectNav challenge at CVPR 2020~\cite{semexp}. It uses a Proximal Policy Optimization method to process a semantic map and target object class, directly providing a policy to sample long-term goals. We used the open-source implementation made available by the authors.
\end{enumerate}
The experimental evaluation used well-established \textbf{measures} for the ObjectNav task in the research literature~\cite{semexp}: 
\begin{enumerate}
    \item \textbf{Success}: ratio of the number of successful episodes to total number of episodes. An episode is successful if the robot is within a fixed distance (1m) of an instance of the target object class.

    \item \textbf{SPL} (Success weighted by path length): measures the efficiency of path taken by robot compared with optimal path; it is is computed as:
    \begin{equation*}
    SPL=\frac{1}{N} \sum_{i=1}^{N} S_i . \frac{l_i}{max(p_i, l_i)}  
    \end{equation*}
    where N is the number of test episodes, $S_i$ is a binary success indicator, $l_i$ is the length of shortest path to closest instance of target object from the robot’s initial position, and $p_i$ is the length of path traversed by robot. 

    \item \textbf{Distance to Success (DTS)}: denotes the distance between the robot and the permissible distance to target for success at the end of an episode.
    \begin{equation*}
    DTS = max( \left \| x_T-G \right \|_2 - d_s, 0 )   
    \end{equation*}
    where $\left \| x_T-G \right \|_2$ is the $L2$ distance between robot and current goal location at the end of the episode; $d_s=1.0 m$ is the success threshold.
\end{enumerate}
% \begin{itemize}
%     \item \textbf{Success:} Ratio of episodes where the method was successful. 
%     \item \textbf{SPL:} Success weighted by Path Length. This metric measures the efficiency of reaching the goal in addition to the success rate.
%     \item \textbf{Distance to Success (DTS):} This is the distance of the agent from the success threshold boundary when the episode ends. This is computed as follows:
%     \begin{equation*}
%         DTS = max(||x_T - G||_2 - d_s, 0)
%     \end{equation*}
%     where $||x_T-G||_2$ is $L2$ distance of the agent from the goal location at the end of the episode, $d_s$ is the success thresold.
% \end{itemize} 
The full list of \textbf{hyperparameters} used in the corresponding algorithms is provided in Table~\ref{hyperparameters}. We kept the settings identical for every experiment of our framework and conducted paired trials to evaluate the different methods (i.e., our framework, baselines) under similar conditions.

%% file: sections/5_results.tex
\section{Experimental Results}
\label{section:results}

Recall that training involved five repetitions of using $1\ million$ frames from $10$ randomly sampled scenes from the benchmark dataset, with each the training models being evaluated in each such repetition on a separate set of five scenes. We conducted $200$ evaluation episodes per scene, leading to a total of 1000 episodes in Gibson (with $5\ scenes$ for evaluation). The corresponding quantitative results are summarized in Table~\ref{table1}. In addition, Table~\ref{table2} summarizes the performance of our framework during each of the five different splits of scenes, considering 10 scenes in each split. We observe that our framework outperformed the baselines; in particular, it achieved a success rate of $63.7\%$ compared with the $57.9\%$ of SemExp~\cite{semexp}, the established state of the art method for the ObjectNav task. 

% Figure~\ref{trajectory} shows an example trajectory obtained when our framework was used to locate an instance of the target object class `potted plant'. The robot's image observations and estimated semantic map are shown. It also depicts the long-term goal selected by the actor network in our framework and the path traversed by the robot to successfully locate an instance of the target object class; blue segmented region in the snapshot in the bottom right of Figure~\ref{trajectory}.

\begin{table}[tb]
\begin{center}
\caption{\textbf{Results} of our framework compared with the baselines, when trained on the train split and evaluated on the val split of the Gibson Tiny dataset. Our framework performs substantially better on all measures.}
\label{table1} 
\begin{tabular}{|c|ccc|} %change to cc for 2 columns
\hline
\multicolumn{1}{|c|}{\textbf{Method}} & \multicolumn{1}{c}{\textbf{Success} }$\uparrow$ & \multicolumn{1}{c}{\textbf{SPL}}$\uparrow$ & \multicolumn{1}{c|}{\textbf{DTS (m)} $\downarrow$}\\
\hline
Random &  0.004 & 0.004 & 3.893\\
SemExp\cite{semexp} &  0.579 & 0.280 & 2.050 \\
\textbf{Our framework} &  \textbf{0.637} & \textbf{0.313} & \textbf{1.568} \\
\hline
\end{tabular}
\end{center}
\end{table}

\begin{table}[tb]
\begin{center}
\caption{\textbf{Results} of our framework in each of the five different splits (of 10 training scenes) from the Gibson dataset.}
\label{table2} 
\begin{tabular}{|c|ccc|} %change to cc for 2 columns
\hline
\multicolumn{1}{|c|}{\textbf{Method}} & \multicolumn{1}{c}{\textbf{Success} }$\uparrow$ & \multicolumn{1}{c}{\textbf{SPL}}$\uparrow$ & \multicolumn{1}{c|}{\textbf{DTS (m)} $\downarrow$}\\

\hline
Split 1 &  0.637 & 0.313 & 1.568 \\
Split 2 &  0.638 & 0.307 & 1.607 \\
Split 2 &  0.616 & 0.301 & 1.777 \\
Split 3 &  0.624 & 0.300 & 1.655 \\
Split 4 &  0.631 & 0.308 & 1.665 \\
\hline
Average & 0.629 & 0.305 & 1.654 \\
\hline
\end{tabular}
\end{center}
\end{table}

\begin{table}[!tb]
\begin{center}
\caption{Performance of our framework with each of the four different \textbf{image augmentation} methods described in Section~\ref{section:approach-semanticmap}; \textit{random shifts} provides the best balance between simplicity and performance.}
\label{table3} 
\begin{tabular}{|c|ccc|} %change to cc for 2 columns
\hline
\multicolumn{1}{|c|}{\textbf{Augmentation}} & \multicolumn{1}{c}{\textbf{Success} }$\uparrow$ & \multicolumn{1}{c}{\textbf{SPL}}$\uparrow$ & \multicolumn{1}{c|}{\textbf{DTS (m)} $\downarrow$}\\
\hline
Flip & 0.597 & 0.272 & 1.845 \\
Grayscale & 0.596 & 0.284 & 1.839 \\
Rotate &  0.604 & 0.284 & 1.706 \\
\textbf{Random Shift} &  \textbf{0.637} & \textbf{0.313} & \textbf{1.568}\\
\hline
\end{tabular}
\end{center}
\end{table}

%\subsection{Ablations}
Recall that we described four different image augmentation methods in Section~\ref{section:approach-semanticmap}. Table~\ref{table3} summarizes the results of evaluating the effectiveness of each of these methods within our framework. These results indicated that the \textit{random shifts} method provides a good balance between simplicity and performance. We thus only used this method for data augmentation in all our experiments, including those summarized in Table~\ref{table1} and Table~\ref{table2}.

% \begin{figure}[tb] 
%         \centering
%         \includegraphics[width=\linewidth]{sections/figures/DTS.png}
%         \caption{\textbf{Quantitative comparison} of our framework with the SemExp baseline based on the DTS measure. The long term goal computed by our method is consistently closer to the nearest instance of target object class than the goal computed by the SemExp baseline.}
%         \label{dts-comparison}
% \end{figure}

% To further evaluate \textbf{H2}, we conducted a comparison of our framework with the SemExp baseline based on the DTS measure; Figure~\ref{dts-comparison} shows the evolution of this measure over a set of episodes. We noticed that our framework enabled the robot to sample and use long-term goals that were consistently closer to the nearest instance of the target object class than the long-term goals computed by the SemExp baseline. In other words, our framework consistently generated better long-term goals for the robot to explore, thus contributing to the observed improvement in performance of the the ObjectNav task. These results establish the importance of learning and using high-level representations from the semantic map, and of incorporating other capabilities (regularization, data augmentation), towards more generalizable and efficient long-term goal selection.

\begin{figure*}[t] 
        \centering
        \includegraphics[width=0.9\textwidth]{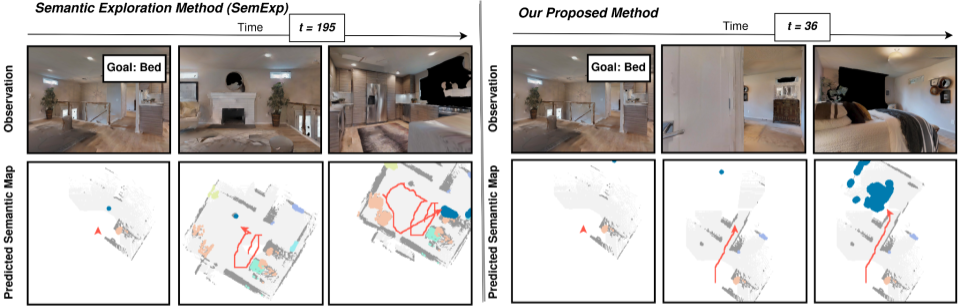}
        \caption{\textbf{Qualitative Comparison with SemExp baseline: } Snapshots from a paired trial comparing our framework (right) with the SemExp baseline~\cite{semexp} (left) in a particular scene. With each method, the robot starts at the same location with the same target object class `bed'. Our framework results in an instance of the target object class being found much faster.}
        \label{comparison}
        \vspace{-1em}
\end{figure*}

%Figure~\ref{comparison} shows a qualitative example of how our framework generates better long-term goals to locate an instance of the target object class much quicker than the SemExp baseline. 

Finally, Figure~\ref{comparison} shows a qualitative comparison between our framework and the SemExp baseline in a particular paired trial in which the robot had to locate an instance of the target object class `bed'. Snapshots from the baseline are provided on the left, and those from our framework are shown on the right. Our framework resulted in the closest instance of the target object class being found much faster, primarily because it generated better long-term goals and supported better generalization. Additional qualitative results and supporting material are available online:\ \\\url{https://user432.github.io/objnav-drq/}

%% file: sections/6_conclusion.tex
\section{Conclusion and Future Work} \label{section:conclusion}
In this paper, we described a modular deep reinforcement learning (DRL) method for the challenging object goal navigation (ObjectNav) task. Our key idea was to treat long-term goal selection (i.e., to determine `where to go')  as a DRL problem, using a combination of an Encoder network and an Actor-Critic network to extract high-level (abstract) representations from an estimated semantic map. In addition, we incorporated simple data augmentation methods and value function regularization methods to improve generalization. Experimental results obtained with the Gibson benchmark dataset in the AI Habitat 3D simulation environment demonstrated that our framework substantially improves performance on standard measures in comparison with a state of the art baseline for the ObjectNav task. Future work will further explore and seek to understand the abstract representations that have contributed to the improved performance. We will also consider other benchmark datasets of scenes for evaluation, and conduct ablation studies to understand the contribution of the actor-critic network to the framework's performance. Furthermore, we will explore the use of the networks trained in simulation on a physical robot operating in an indoor environment.